\documentclass{article}

\usepackage[final]{nips_2016}

\usepackage[utf8]{inputenc} % allow utf-8 input
\usepackage[T1]{fontenc}    % use 8-bit T1 fonts
\usepackage{hyperref}       % hyperlinks
\usepackage{url}            % simple URL typesetting
\usepackage{booktabs}       % professional-quality tables
\usepackage{amsfonts}       % blackboard math symbols
\usepackage{amsmath}
\usepackage{nicefrac}       % compact symbols for 1/2, etc.
\usepackage{microtype}      % microtypography
\usepackage{mwe}
\usepackage{graphicx}
\usepackage{tikz}
\usepackage{float}

\usepackage{amsmath}
\usepackage[linesnumbered,ruled]{algorithm2e}

\usepackage{mathtools, graphicx,amsmath,amsthm, amssymb,epstopdf,verbatim,array,booktabs,mathtools}  

\title{BlockPuzzle — A Challenge in Physical Reasoning and Generalization for Robot Learning}

\author{
  Yixiu Zhao\\
  School of Computer Science\\
  Carnegie Mellon University\\
  Pittsburgh, PA 15213 \\
  \texttt{yixiuz@andrew.cmu.edu} \\
  \And
  Ziyin Liu\\
  Department of Mathematical Sciences\\
  Carnegie Mellon University\\
  Pittsburgh, PA 15213 \\
  \texttt{ziyinl@andrew.cmu.edu} \\
}

\begin{document} %

\maketitle

\begin{abstract}
In this work we propose a novel task framework under which a variety of physical reasoning puzzles can be constructed using very simple rules. Under sparse reward settings, most of these tasks can be very challenging for a reinforcement learning agent to learn. We build several simple environments with this task framework in Mujoco and OpenAI gym and attempt to solve them. We are able to solve the environments by designing curricula to guide the agent in learning and using imitation learning methods to transfer knowledge from a simpler environment. This is only a first step for the task framework, and further research on how to solve the harder tasks and transfer knowledge between tasks is needed.
\end{abstract}

\section{Introduction}
Generalizability is one of the most important problems in research for artificial general intelligence. Ideally, we want our algorithms to be able to generalize to unseen circumstances. In the context of reinforcement learning, we hope that our agents can master games in such way that when the objects are spawned in a different configuration, they can still play the game. An even more challenging direction is task transfer, where the agent can adapt to new games with similar rules with little or no training. 

In this work, we propose a novel task framework in which a variety of different tasks can be constructed under the same set of simple rules. This is the first step towards a ``generalizable reinforcement learning". This might be rephrased into a problem of transfer learning: how do trained models behave in the planning for unseen circumstances? What is the cause of its generalization or its inability to generalize? In context of physical problem solving, generalization clearly requires a certain level of scene understanding and intuitive understanding of physics: it has been argued that data-driven approaches that perform pattern recognition might fail to generalize beyond training data and therefore is different from human learning \cite{tenenbaum2011grow}. Therefore, the ultimate goal of artificial intelligence is a learning framework for agents in simulated environments that should be able generalize to unseen scenarios well with no or minimal amounts of additional training. 

Our main contribution is threefold: 
\begin{itemize}
    \item A principle for testing generalizability of artificial agents outside the training environment it is trained on
    \item A (series of) environment of increasing difficulty that tests the capability of the agent to generalize
    \item A collection of baselines on this environment that either succeeds in or fails to learn the task
\end{itemize}

\section{Related Works}
\subsection{OpenAI Robotics Environments}
Learning control schemes is a important and practical topic in Reinforcement Learning. However, most Reinforcement Learning algorithms are extremely sample inefficient. If trained in the real world with these methods, a robot would have to fail millions of times to know what the right things to do is. Therefore, it is greatly beneficial to train the model on a simulator and then transfer the policy onto the real world. For this purpose, OpenAI released a series of robotics control environments \cite{gym} based on the physics simulation engine Mujoco \cite{mujoco}.

The environments involve two types of robots: \textit{Fetch}, a robotic arm with 7 degrees of freedom, and \textit{ShadowHand}, a robotic hand with 20 degrees of freedom. The Fetch environments consists of tasks such as reaching a certain position in space, pushing a block to a certain position on the table, and sliding a puck to a position where the robot arm cannot reach. The ShadowHand environments involve orienting objects with various shapes to a desired orientation in the hand. 

All of these environments provided by OpenAI are goal-oriented, which is to say that there is a determined goal the success rate on which the agents are evaluated on. Particularly, these goals can be expresses in a simple vector, which can be compared with the vector corresponding to the current state to determine if the goal is reached. For example, for the \textit{FetchReach-v0} environment, the goal is the target position in space that the robot is trying to reach, and the current reached goal is the position of the robot's grippers. In the sparse reward setting of the environments, a reward is given only when the goal is reached, and in the dense reward setting, rewards are given according to the distance between the desired goal and current goal reached. This formulation of goals makes possible the Hindsight Experience Replay (HER) method \cite{andrychowicz2017hindsight}, which treats failure episodes of the agents as potentially successful episodes with a different goal specified. This greatly benefits training since the reward signals are increased greatly for these environment.

%Despite the success of HER, we must be reminded that most real world tasks have more complicated structure and cannot be formulated by this framework. Consider an autonomous driving system, which not only has a goal of reaching the destination, but also needs to satisfy constrains such as not hitting any pedestrians along the way. Also, goals can be more abstract and allow multiple states in which it is considered complete. Consider the task of building a tower of blocks. There is no restriction on which blocks are on tops, so the goal allows multiple states, and it is hard to find a simple and useful formulation of the goal vector to give information on how close the agent is to success. In Section 3, we present a different task framework and addresses this issue and a series of environments under this new framework based on the OpenAI robotics environments.

\subsection{Curriculum Learning}
Of the many methods to guide the agent into the desired behavior, curriculum learning is one of the most general and successful frameworks \cite{Curriculum}. It is also the most intuitive since it is how humans learn most subjects in school. For our environments, we choose the curriculum learning framework where the task is fixed and the distribution of the starting state varies \cite{reverse_curriculum}. Let $\rho_0:\mathbb{S}\rightarrow\mathbb{R}_+$ be the distribution of the start state that we evaluate the agent on. In training, we use different distributions $\rho_i$ such that it is easier for the agent to get reward signals to learn useful information. Once the agent reaches a certain level of performance on distribution $\rho_i$, we switch to the next distribution $\rho_{i+1}$. In the case where the hardness of $\rho_i$ increases smoothly and converges to $\rho_0$, we expect the agent to be able to learn to perform the task well on the test distribution $\rho_0$.

\subsection{Imitation Learning: AggreVaTeD}
In some tasks that are sufficiently difficult, we perform imitation learning. However, in some cases we do not have a near optimal policy teacher for our network to learn completely through its demonstration, in this case, a mixed imitation learning and reinforcement learning is needed \cite{aggravated}. In the best case, the AggreVaTeD algorithm can provide up to exponential lower sample complexity than pure reinforcement leaning. While the theory is very involved, the algorithm is simple to state. We first define a teacher-forcing ratio $a$, according to which our agent either gets trained by supervision learning on the output of the teacher, or by the reward signal from reinforcement learning. It is expected that the teacher offers the most help early in the training and not so much late in the training, and so the algorithm anneals the teacher forcing ratio from $a_{max}$ to $a_{low}$ through some predefined scheduling function. In short, an unbiased and variance reduced of the loss gradient is:

\begin{equation}
    \hat{\nabla_{\theta_n}} = \frac{1}{HK} \sum^K_{i=1}\sum^H_{t=1}\frac{\nabla_{\theta_n} \pi_{\theta_n} (a_t^{i,n}|s_t^{i,n})}{\pi_{\theta_n}(a_t^{i,n}|s_t^{i,n})} A^*_t(a_t^{i,n}|s_t^{i,n})
\end{equation}
where we have used importance sampling because the action space is continuous. For more detail, see \cite{aggravated}. The algorithm is very simple:

\begin{algorithm}
    \SetKwInOut{Input}{Input}
    \Input{The given MDP and expert $\pi^*$, learning rate $\eta_n$, schedule rate $a_i$, where $a_n \to 0$ as $n \to \infty$}

     Initialize policy $\pi_{\theta_i}$\;
     
    \For{$n=1$ to N}
    {
              Mixing policies: $\hat{\pi}_n = a_n\pi^* + (1-a_n)\pi_{\theta_n}$\; 
              Starting from $\rho_0$, roll in by executing $\hat{\pi}_n$ on the given MDP to generate $K$ trajectories $\{\tau_i^n\}$ \;
              Using $Q^*$ and $\{\tau_i^n\}_i$, compute the descent direction $\delta_{\theta_n}$\;
              $\theta_{n+1}=\theta_n - \eta_n\delta_{\theta_n}$ \;
    }
    \Return the best hypothesis $\{\hat{\pi}_n\}$ on validation\;

    \caption{Differentiable AggreVaTe}
\end{algorithm}

\subsection{Deep Deterministic Policy Gradients}
Deep Deterministic Policy Gradients (DDPG) is a policy gradient algorithm that uses a stochastic behavior policy for good exploration but estimates a deterministic target policy, which is much easier to learn \cite{lillicrap2015continuous}. DDPG is also based on actor-critic algorithms; it primarily uses two neural networks, one for the actor and the other the critic. These networks computes action predictions for the current state and generate a temporal difference error signal each time step. The input of the actor network is the current state, and the output is a single real value representing an action chosen from a continuous action space. The loss function is:
\begin{equation}
    L = \frac{1}{N}\sum_i (y_i - Q(s_i, a_i| \theta^Q)^2)
\end{equation}

differentiating with respect to this gives us the update rule for the actor network, and the the stochastic version of it is \cite{Silver:2014:DPG:3044805.3044850}:
\begin{equation}\label{ddpg}
    \nabla_{\theta^\mu} \mu \approx \mathrm{E}[\nabla_a Q(s, a | \theta^Q) |_{s=s_t, a=\mu(s_t)} \nabla_{\theta^\mu} \mu(s|\theta^\mu )|_{s=s_t}]
\end{equation}
In fact, this is true as long as the Markov decision process satisfies some appropriate conditions, for more detail see \cite{Silver:2014:DPG:3044805.3044850}. This tells us that the stochastic policy gradient is equivalent to the deterministic policy gradient. The pseudo-code for DDPG is:

\begin{algorithm}

    \SetKwInOut{Input}{Input}
    Randomly initialize critic network $Q(s,a|\theta^Q)$ and actor $\mu(s|\theta^\mu)$ with weights $\theta^Q$ and $\theta^\mu$ \;
    Initialize target network $Q'$ and $\mu'$ with weights $\theta^{Q'} = \theta^Q$, $\theta^{\mu'} = \theta^\mu$ \;
    Initialize replay buffer $R$\;

    \For{episode = 1 to M}
    {
              Initialize a random process $G$ for action exploration\;
              Receive initial observation state $s_1$\;
             
              \For{t=1 to T}{
                  Select action $a_t = \mu(s_t|\theta^\mu) + \mathcal{N}_t$ according to the current policy and exploration noise\;
                  Execute action $a_t$ and observe reward $r_t$ and observe new state $s_{t+1}$\;
                  Store transition $(s_t, a_t, r_t, s_{t+1})$ in $R$\;
                  Sample a random minibatch of $N$ transitions $(s_i, a_i, r_i, s_{i+1})$ from $R$\;
                  Set $y_i = r_i + \gamma Q'(s_{i+1}, \mu'(s_{i+1}| \theta^{\mu'}) | \theta^{Q'})$\;
                  Update critic by minimizing the loss: $L=\frac{1}{N}\sum_i(y_i-Q(s_i, a_i|\theta^Q))^2$\;
                  Update the actor policy using the sampled policy gradient according to eq.~\ref{ddpg}\;
                  
                  Update the target networks:
                  \[\theta^{Q'} = \tau \theta^Q + (1 -\tau)\theta^{Q'}\]
                  \[\theta^{\mu'} = \tau \theta^Q + (1 -\tau)\theta^{\mu'}\]
                }

    }
    \caption{Deep Deterministic Policy Gradient}
\end{algorithm}

\section{Environment}

\begin{figure}
    \centering
    \includegraphics[width=0.99\textwidth]{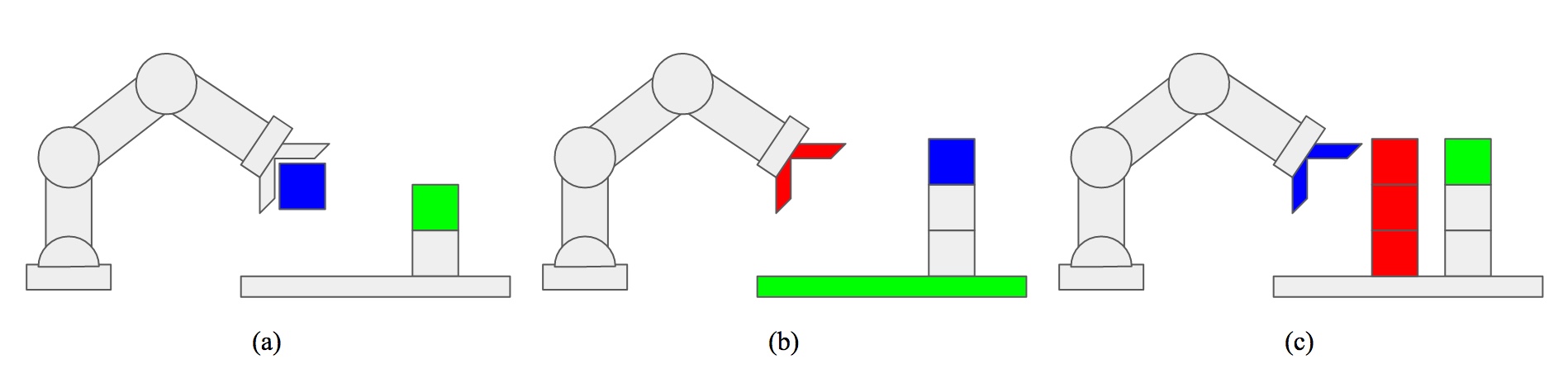}
    \caption{Illustration of the designed task. In each case the robot aims to put the green and blue object into contact, under the constraint that the red and blue object should not touch.}
    \label{fig:env}
\end{figure}

% \subsection{The Principle}
% In this section we describe our environment (fig.~\ref{fig:env}). The environment is build in the MuJoCo environment \cite{mujoco}, and is based on the robotics tasks in the OpenAI gym environment \cite{gym}, in this environment, the agent is a robotics arm with 7 degrees of freedom and each control consists of a target position and orientation to linearly move the clamp to, and the hold/release of the clamp. This is a continuous control problem, and is trainable with algorithms such as DDPG \cite{DDPG}. At the end of each iteration the agent will receive a reward based on the task framework, the cost of executing the action and the cost for consulting the experts. While the OpenAI robotics task gives a few tasks on basic operations, we find them insufficient for the testing of high level scene and task understanding. Towards improving this, we first propose three criterion for the tasks for training testing a generalizable agent:

% \begin{itemize}
%     \item the agent should learn about the correct goal of the task and learn about the efficient ways to achieve it, either directly, or indirectly.
%     \item the existence of noise objects or features that can be effective identified as irrelevant should not hinder the performance of the agent
%     \item where there are external constraints or rules, the agents should find an efficient way to achieve the task without violating the rule
% \end{itemize}

\subsection{Task Framework}
Based on the discussion before, we propose a novel framework consisting of simple rules that supports a wide range of robotics tasks. In the framework objects are colored with four colors: red, blue, green and grey. When a red (constraint) object contacts with a blue (manipulation) object, the task is considered failed. When all blue objects are in contact with a green (goal) object at some point, the task is considered complete. Grey objects are neutral and does not contribute to the success condition when touching other objects. Also, nothing special happens when a red object contacts with a green object. Note that the rules for different colored blocks touching can be made more general, and we are only considering the simplest set from which interesting tasks can be constructed.

Under the color scheme in the aforementioned framework, consider the above situations (fig.~\ref{fig:env}). The first situation corresponds to a block contact problem, which, with more constraints specified, can become a problem for block stacking. The second one corresponds to a task of toppling the block tower so that the blue block falls onto the green table. The third situation corresponds to a path planning problem where the robot arm has to reach the green block while avoiding the obstacles. These are just simple examples of what tasks the framework is able to cover, and there can be many more complicated variations and combinations of these tasks. In fact, similar versions of many previously studied robotics tasks such as fetching, pushing and stacking blocks, along with more sophisticated tasks such as path planning with obstacles and toppling a block tower in a certain direction can all be implemented in this framework.

Note that different arrangement of blocks and colors such as shown in the figure are considered to be different tasks. However, since they are under the same set of rules, it is natural to assume that if the agent learns how these simple rules work as well as skills for manipulating blocks from the tasks during training, it should be able to generalize to other unseen tasks and solve them as well. In fact, it is obvious that humans with basic motor skills, scene understanding and intuition for physics should be able to solve many of these different unseen tasks with ease. We hypothesize that current RL methods still rely on large numbers of training scenarios and will overfit the environments they are trained on. As a consequence they will fail to generalize to unseen types of tasks under our framework. We think that generalization or transfer learning performance on different tasks within this framework can be very challenging and is a good measure of a system's scene understanding and intuitive physics capabilities. In this paper, we attempt to kickstart the process towards fully solving the task framework by performing existing methods on two simple environments under the framework.

\subsection{Tested Environments}
Here we describe our environments \textit{BlocksTouch-v0} and \textit{BlocksChoose-v0}, which we performed experiments on. The environments are built in the MuJoCo environment \cite{mujoco}, and are based on the robotics tasks in the OpenAI gym environment \cite{gym}. In the OpenAI Fetch robotics environments, the agent is a robotics arm with 7 degrees of freedom with a clamp to pickup objects. However for the purpose of our experiments, several degrees of freedom, including control of the gripper, are locked, and the agent only have to output a 4-dimensional action. The actions $\mathcal{A}\in \mathbb{R}^4$ are real-valued torques applied to the joints of the robot, and each is normalized to $-1$ and $1$. The observations contains the position, velocity and gripper states of the robot, as well as the position, orientation, velocities and color of each block, and concatenated sequentially.

The \textit{BlocksTouch-v0} environment has two blocks, a green one and a blue one, which need to come into contact, while in the environment \textit{BlocksChoose-v0} there is an extra block to interfere with the agent. To lower the hardness of the task, we always put the grey block in the last few dimensions in the observation. Screenshots taken from running of our environment is given in fig.~\ref{fig:video}.

%\begin{figure}
%    \centering
%    \includegraphics[width=175pt]{plots/demo1.png}
%    \includegraphics[width=180pt]{plots/demo2.png}
%    \caption{Real time running of our environment.}
%    \label{fig:demo}
%\end{figure}

% \subsection{Training Regimes}

% During some preliminary experiments we realized that the tasks, despite their simple-looking disguise, can be very hard for a zero-knowledge agent to be trained on. We thus introduce two more optional auxiliary training regimes into our environment. We first introduce a curriculum training regime.

\subsection{Curriculum Settings}
% Curriculum training is a training regime in which the difficulty of the tasks are gradually increased for an agent, and the key assumption behind the working of curriculum training is that it is easier for the agent to learn to ``extrapolate" from easier tasks to harder tasks than to learn directly from hard environments. Curriculum training \cite{Curriculum} has proved to be effective at training deep neural networks.

As an example, the 2 block task is implemented in the following way. We first fix the arm to start at a default position for every episode. Now define a maximum radius $R$, such that the first block appears within radius $R$ to the arm uniform randomly; we then sample the second block, which appears within radius $R$ to the first block uniformly randomly. This finishes the set up of the episode, and we start the episode from here. We start from a very small $R$ and gradually increase it to include the whole table at the highest difficulty (while making sure that the blocks are not sampled outside the table). For the 3 blocks case, we also include a minimum radius $R_{min}$, the distance from the center point between the colored blocks under which the grey block would not spawn. This value starts high and is gradually decreased to 0. We also define a level threshold $h$, and for every epoch, if the training success rates pass the threshold $h$, then we increase the difficulty to the next level. In practice, we find $h=0.7$ tend to work well. We noticed that curriculum training greatly increased the training speed and convergence rate of the baseline agents. A 3-layer baseline model takes fewer than 50 epochs to converge in this training regime.

\section{Methods}
\subsection{Policy Gradient Methods}
In the experiments we tried two variants of Policy Gradient: Deep Deterministic Policy Gradient (DDPG) provided by OpenAI \cite{gym}, and our own variant which produces a normal distribution over the space of continuous actions, which we call Policy Gradient with Gaussian Distribution (PGGD). The OpenAI implementation of DDPG uses the actor-critic scheme. The actor's objective is simply the negative value function $-Q(s,a)$ predicted by the critic, while the critic's objective is the commonly used TD-learning objective. The inclusion of the critic reduces the variance and makes it more stable in training, but it also complicates the implementation of imitation learning, since both the actor and the critic needs to be learning from the expert. By contrast, PGGD does not have a critic model, and the actor produces a Gaussian distribution over the action space, from which one action is sampled and executed. This can be represented as $a(s, \theta)\sim\mathcal{N}(\mu(s,\theta), \sigma^2(s, \theta))$. The update rule for PGGD is the same as in normal policy gradient. Here is a the two algorithms in pseudocode:

Note that during training, the variance of the distribution produced by PGGD will never decrease to 0, and while the stochasticity is good for exploration, it is not good for evaluating the performance of the algorithm. Therefore we divide performance measure into three categories: training, testing and finals. In testing and finals, we use the mean of the Gaussian for a more stable evaluation of the performance. In training and testing, the evaluations are performed on the current level of curriculum, and the success rate in testing will determine if the agent is ready for the next level of difficulty. The finals is evaluated on the maximum difficulty and reflects the true training progress.

\subsection{Imitation Learning}
In our experiments we found that the 3-blocks environment is significantly harder than the 2-blocks environment, even with the application of curriculum learning. Since these task are similar in structure, and the third block mainly serves as a distraction and hindrance in completing the task, we wish to transfer useful knowledge of the learned agent from the 2-blocks case to the 3-blocks environment.

We used the AggreVaTeD framework to transfer knowledge between the expert and the agent being trained. The expert used in our experiments is a DDPG policy trained on the 2 blocks environment, and the learner policy is a PGGD policy in the 3-blocks environment. During the training of the learner policy, expert take control of the robot with a probability $\beta$ with is annealed exponentially:

$$\beta=\beta_0+(1-\beta_0)e^{-\frac{t}{t_0}}$$

Where $t_0$ is controls the rate of annealing and $\beta_0$ controls the independence of the learner policy. Whenever the expert is in control, the state from the environment is processed so that the grey block is removed from the observation. So the expert performs actions while not seeing the grey block. This scheme produces experiences with good reward signals since the grey block is far away from the colored ones and do not interfere with the task early on in the curriculum. However, as the curriculum gets harder, there is an increasing chance that the grey block is spawned in between the colored blocks, making the expert fail on completing the task.

\section{Experiments}
The hyperparameters we used for DDPG is mostly identical to those reported in \cite{plappert2018multi}. The actor and critic networks are both MLPs with 3 layers and 256 hidden units each, with ReLU activation \cite{Nair_relu}. The input is the normalized state observations. Tanh activation of the output is used for the actor to produce a valid action. We tried some variants in network depth and learning rate, and found the original ones to be the optimal. For PGGD we used linear activation for the mean and softplus for the standard deviation, and the learning rate is $0.0001$. Both policies are trained off-policy with experience replay \cite{Lin1992}, with a batch size of $256$. For imitation learning we used $\beta_0=0$ and $t_0=50$, which we empirically found to yield good results.

Each epoch of training consists of 50 cycles of training, in which every MPI worker records rollouts in the replay memory and trains on 40 batches sampled from the memory. All experiments are trained to a maximum of 200 epochs.

We originally performed our experiments on an AWS machine with 2 cores, using 2 rollouts per MPI worker. We discovered that the number of workers used does not impact the learning of PGGD significantly. However, when we ran experiments on another machine with 20 cores, we discovered a significant boost of performance on DDPG, yielding more competitive results learning from scratch than more sophisticated methods like PGGD+AggreVaTeD on the 3-blocks environment. Despite this fact, it is still true that PGGD+AggreVaTeD outperforms vanilla DDPG under low sample size constraint, and therefore can be more sample efficient under these circumstances.

\subsection{Baselines}

We found the 2 blocks environment to be fairly easy to learn for DDPG. The policy reached 70\% accuracy after 75 epochs while trained on 2 cores, and it reached 95\% after 20 epochs while trained on 20 cores (fig.~\ref{fig:plots}, top left). Watching the behaviour of the agents, we found that the agent learned to always push one of the two blocks towards the other one, instead of reaching for the closest one. This sometimes leads to failure when one the blocks is a little too far away from the robot for it to retrieve. This suggests that although a high level of performance is reached, the agent still has no concept that both blocks can be moved to achieve the same goal.

The baseline for 3 blocks is much harder to train on 2 cores for DDPG, as the algorithm can not even get past the first difficulty level, where the grey block is position far away from the colored blocks so that it does not interfere at all. Tweaking with the spawning position of the third block, we discover that even the tiniest variation of the spawning position of the third block produces great fluctuations in training. This is likely due to the fact that without enough experience as evidence, the network does not know that the grey block is irrelevant and should be ignored at this point. We tried changing our curriculum so that the algorithm can eventually learn this, but it turned out to be too slow to be meaningful.

\subsection{Imitation Learning}
Here we cover the results obtained with PGGD+AggreVaTeD, with the agent trained in the two blocks environment in the previous section as the expert. We notice that with the expert policy as guidance, the learner quickly picks up on what the right thing to do is, passing the lower difficulty levels with ease. Eventually the training reaches 67\% success rate at around 250 epochs (fig.~\ref{fig:plots}, top right). Note that this is much better than training PGGD alone with the same hyperparameters, where the agent barely passes the first difficulty level after 200 epochs. The performance is also better than that of the expert, with a success rate of merely 44\% due to the colored blocks being further apart and the grey block getting in the way from time to time. Also note that although the data is produced with 20 cores, the similar level of performance can be reached with only 2 cores. These facts show that PGGD+AggreVaTeD is very effective at kickstarting training and obtaining higher performance by learning from an imperfect expert.

\subsection{DDPG with 20 Cores}
\begin{figure}[H]
    \centering
    \includegraphics[width=175pt]{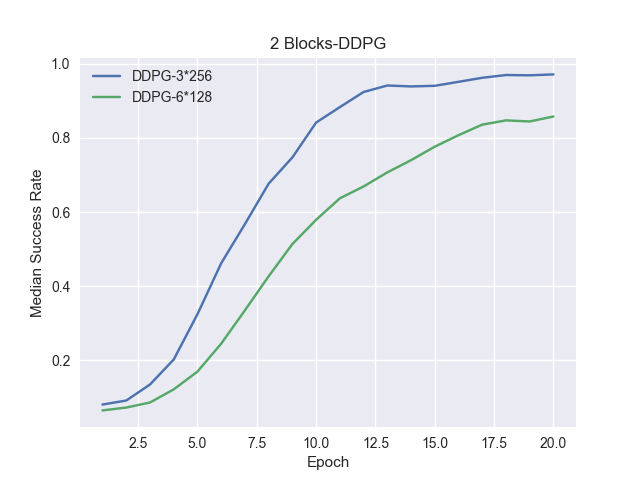}
    \includegraphics[width=180pt]{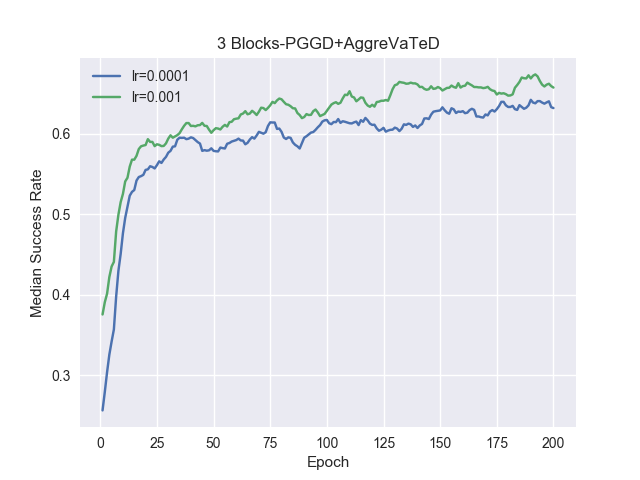}
    \includegraphics[width=175pt]{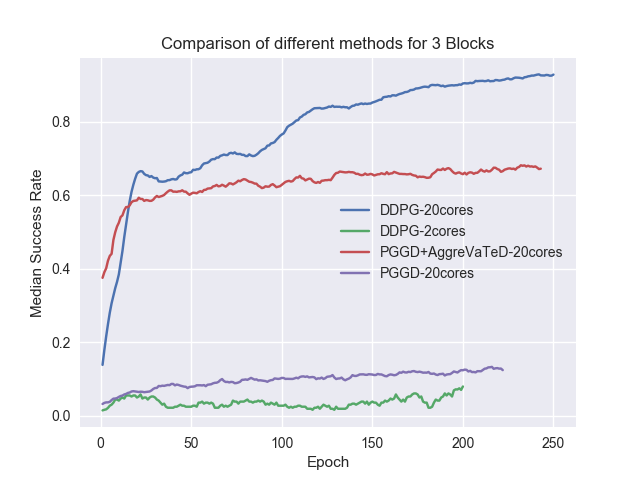}
    \includegraphics[width=180pt]{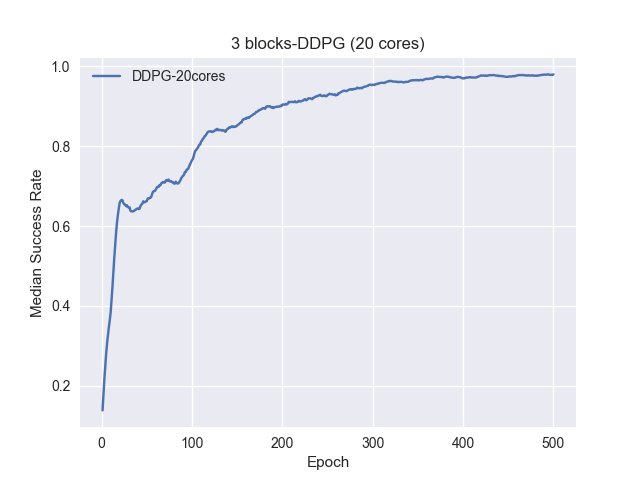}
    \caption{Training plots of our experiments on the two environments.}
    \label{fig:plots}
\end{figure}

\begin{figure}[H]
    \centering
    \includegraphics[width=350pt]{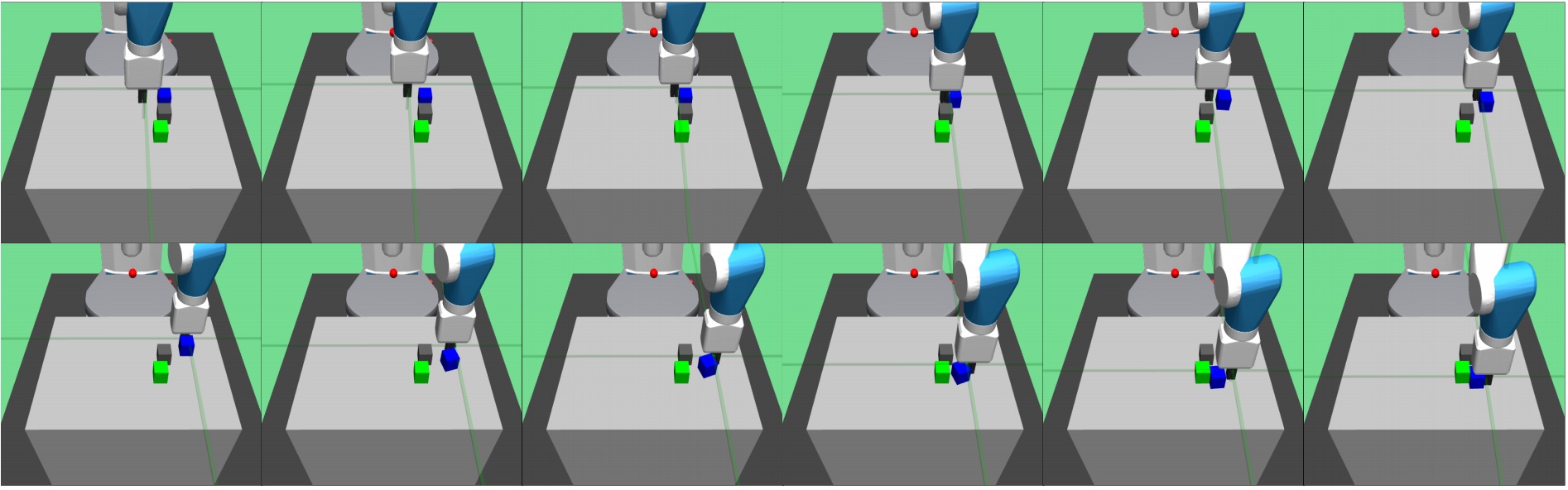}
    \includegraphics[width=350pt]{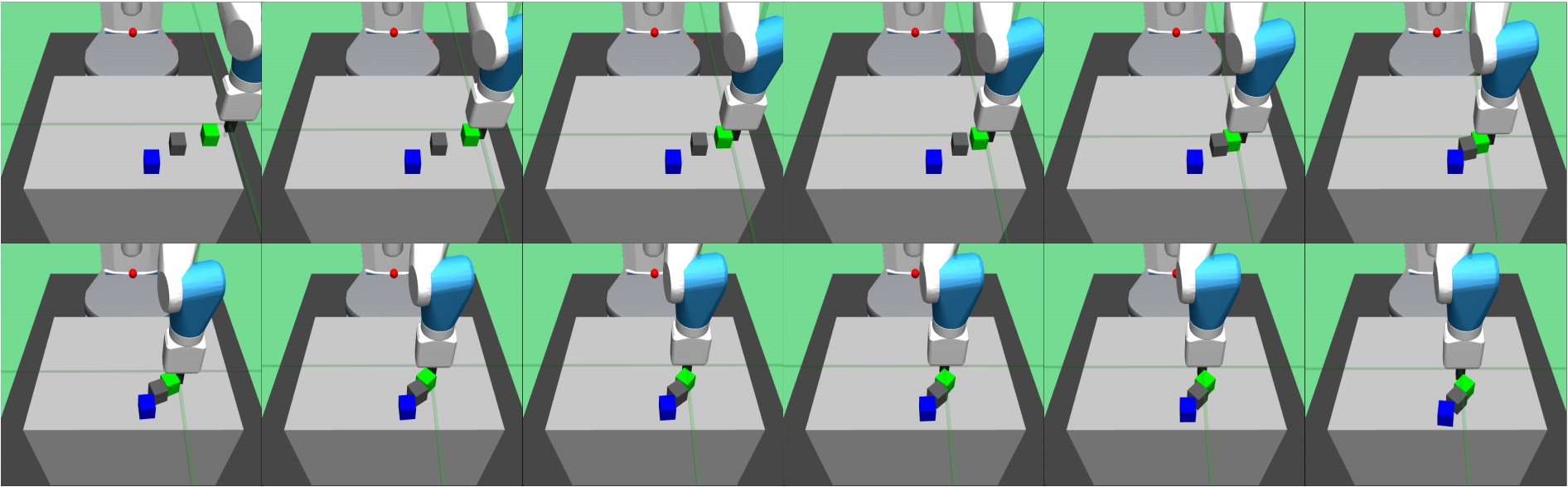}
    \caption{Examples of the agent dealing with the grey block. 
    Top: The agent learns to push the blue block around the grey block; Bottom: The agent does not avoid the grey block, leading to a failure.}
    \label{fig:video}
\end{figure}

During our final run of the experiment, we discovered the surprising fact that DDPG with 20 cores actually outperforms PGGD+AggreVaTeD by a large margin, with a success rate of 90\% at 200 epochs, reaching a summit of 97\% at 500 epochs (fig.~\ref{fig:plots}, bottom left/right). We suspect that this is the result of having a critic which can learn from a more independently distributed set of data to guide the actor. In particular, we hypothesized that DDPG was able to learn to avoid the third block by moving other blocks around it. To test this hypothesis, we constructed a final challenge level for the agents, where the colored blocks are at least 0.15 distance apart, and the grey block spawns at the center point of the two colored blocks. We observed that PGGD+AggreVaTeD does not know to avoid the grey block, and can just end up pushing all three blocks off the table, while DDPG learned to maneuver around the grey block almost every single time (fig.~\ref{fig:video}). Note that this configuration of blocks is rare even for the highest difficulty level in the training scenarios. This indicates that the agent is already capable to generalize to a different distribution of test cases, indicating a minimal understanding of the physics of this block puzzle. The final results are summarized in Table.~\ref{table:results}.

\newpage

\begin{table}[H]
\label{sample-table}
\begin{center}
\begin{tabular}{l|ll}
\multicolumn{1}{c}{\bf Method}  &\multicolumn{1}{c}{\bf Normal} &\multicolumn{1}{c}{\bf Challenge}
\\ \hline
DDPG(2 blocks)       &0.44      &0.24\\
\textbf{DDPG(3 blocks)}      &\textbf{0.97}       &\textbf{0.90}\\
PGGD+AggreVaTeD     &0.64       &0.40\\
\end{tabular}
\end{center}
\caption{Final evaluation of different agents on the 3 blocks environment.}
\label{table:results}
\end{table}
% \newdimen\X
% \begin{tikzpicture}[
%           picture/.style={draw,rectangle,anchor=#1,inner sep=0pt,outer sep=0pt}
%      ]
%      \X=0mm
%      \foreach \Image in \SuccessOne {
%       \node[picture=north west] at (\X,-76.5mm)
%              {\includegraphics[width=27mm,height=25mm]{\Image}};
%       \global\advance\X by 27mm
%      }
%      \X=0mm
%      \foreach \Image in \SuccessTwo {
%       \node[picture=north west] at (\X,-101.5mm)
%              {\includegraphics[width=27mm,height=25mm]{\Image}};
%       \global\advance\X by 27mm
%      }
     
%           \X=0mm
%      \foreach \Image in \FailOne {
%       \node[picture=north west] at (\X,-150.5mm)
%              {\includegraphics[width=27mm,height=25mm]{\Image}};
%       \global\advance\X by 27mm
%      }
%      \X=0mm
%      \foreach \Image in \FailTwo {
%       \node[picture=north west] at (\X,-175.5mm)
%              {\includegraphics[width=27mm,height=25mm]{\Image}};
%       \global\advance\X by 27mm
%      }
%   \end{tikzpicture}

\section{Discussion}
In this paper, we propose a novel task framework under which a variety of tasks can be formalized. We constructed two simple environments in Mujoco and successfully solved them with the help of curriculum learning and imitation learning. However, a lot still remains to be done. The two environments are simplified for the purpose of the experiment, and the task would be much harder if in the 3 blocks environment the colors are shuffled instead of fixed. A possible next step can also be building a more sophisticated network to handle any possible number of blocks.

We believe that the proposed environment is a novel framework to assess whether a learning agent has an understanding of physical reasoning, as opposed to mere pattern matching. In the future, we wish to algorithms that can not only solve one of the tasks with sparse rewards, but also use that knowledge and understanding of the task structure to transfer to other tasks in the framework.

\clearpage
\bibliographystyle{alpha}
\bibliography{main}

\newcommand{\etalchar}[1]{$^{#1}$}
\begin{thebibliography}{PAR{\etalchar{+}}18b}

\bibitem[ACR{\etalchar{+}}17]{andrychowicz2017hindsight}
Marcin Andrychowicz, Dwight Crow, Alex Ray, Jonas Schneider, Rachel Fong, Peter
  Welinder, Bob McGrew, Josh Tobin, OpenAI~Pieter Abbeel, and Wojciech Zaremba.
\newblock Hindsight experience replay.
\newblock In {\em Advances in Neural Information Processing Systems}, pages
  5055--5065, 2017.

\bibitem[BLCW09]{Curriculum}
Yoshua Bengio, J{\'e}r\^{o}me Louradour, Ronan Collobert, and Jason Weston.
\newblock Curriculum learning.
\newblock In {\em Proceedings of the 26th Annual International Conference on
  Machine Learning}, ICML '09, pages 41--48, New York, NY, USA, 2009. ACM.

\bibitem[FHW{\etalchar{+}}17]{reverse_curriculum}
C.~{Florensa}, D.~{Held}, M.~{Wulfmeier}, M.~{Zhang}, and P.~{Abbeel}.
\newblock {Reverse Curriculum Generation for Reinforcement Learning}.
\newblock {\em ArXiv e-prints}, July 2017.

\bibitem[LHP{\etalchar{+}}15]{lillicrap2015continuous}
Timothy~P Lillicrap, Jonathan~J Hunt, Alexander Pritzel, Nicolas Heess, Tom
  Erez, Yuval Tassa, David Silver, and Daan Wierstra.
\newblock Continuous control with deep reinforcement learning.
\newblock {\em arXiv preprint arXiv:1509.02971}, 2015.

\bibitem[Lin92]{Lin1992}
Long-Ji Lin.
\newblock Self-improving reactive agents based on reinforcement learning,
  planning and teaching.
\newblock {\em Machine Learning}, 8(3):293--321, May 1992.

\bibitem[NH10]{Nair_relu}
Vinod Nair and Geoffrey~E. Hinton.
\newblock Rectified linear units improve restricted boltzmann machines.
\newblock In {\em Proceedings of the 27th International Conference on
  International Conference on Machine Learning}, ICML'10, pages 807--814, USA,
  2010. Omnipress.

\bibitem[PAR{\etalchar{+}}18a]{gym}
M.~{Plappert}, M.~{Andrychowicz}, A.~{Ray}, B.~{McGrew}, B.~{Baker},
  G.~{Powell}, J.~{Schneider}, J.~{Tobin}, M.~{Chociej}, P.~{Welinder},
  V.~{Kumar}, and W.~{Zaremba}.
\newblock {Multi-Goal Reinforcement Learning: Challenging Robotics Environments
  and Request for Research}.
\newblock {\em ArXiv e-prints}, February 2018.

\bibitem[PAR{\etalchar{+}}18b]{plappert2018multi}
Matthias Plappert, Marcin Andrychowicz, Alex Ray, Bob McGrew, Bowen Baker,
  Glenn Powell, Jonas Schneider, Josh Tobin, Maciek Chociej, Peter Welinder,
  et~al.
\newblock Multi-goal reinforcement learning: Challenging robotics environments
  and request for research.
\newblock {\em arXiv preprint arXiv:1802.09464}, 2018.

\bibitem[SLH{\etalchar{+}}14]{Silver:2014:DPG:3044805.3044850}
David Silver, Guy Lever, Nicolas Heess, Thomas Degris, Daan Wierstra, and
  Martin Riedmiller.
\newblock Deterministic policy gradient algorithms.
\newblock In {\em Proceedings of the 31st International Conference on
  International Conference on Machine Learning - Volume 32}, ICML'14, pages
  I--387--I--395. JMLR.org, 2014.

\bibitem[SVG{\etalchar{+}}17]{aggravated}
W.~{Sun}, A.~{Venkatraman}, G.~J. {Gordon}, B.~{Boots}, and J.~A. {Bagnell}.
\newblock {Deeply AggreVaTeD: Differentiable Imitation Learning for Sequential
  Prediction}.
\newblock {\em ArXiv e-prints}, March 2017.

\bibitem[TET12]{mujoco}
E.~Todorov, T.~Erez, and Y.~Tassa.
\newblock Mujoco: A physics engine for model-based control.
\newblock In {\em 2012 IEEE/RSJ International Conference on Intelligent Robots
  and Systems}, pages 5026--5033, Oct 2012.

\bibitem[TKGG11]{tenenbaum2011grow}
Joshua~B Tenenbaum, Charles Kemp, Thomas~L Griffiths, and Noah~D Goodman.
\newblock How to grow a mind: Statistics, structure, and abstraction.
\newblock {\em science}, 331(6022):1279--1285, 2011.

\end{thebibliography}

\medskip
\small
% [1] Bengio, Y. (2009) Learning Deep Architectures for AI.  {\it Foundations Vol. 2}, pp.\ 1--50.

% [2] Shwartz-Ziv, R.\ \& Tishby, N.\ (2017) Opening the black box of deep neural networks via information. {\it arXiv preprint arXiv:1703.00810.}

% [3] Tishby, N., Pereira F. C.\ \& Bialek, W.\ (1999) The Information Bottleneck Method. {\it Proceedings of 37th Annual Allerton Conference on Communication, Control and Computing:} pp.\ 368--377.

% [4] Srivastava, N., Mansimov, E.\ \& Salakhutdinov, R.\ (2015) Unsupervised learning of video representations using LSTMs. {\it International Conference on Machine Learning.}

\end{document}